\documentclass[letterpaper, 10 pt, journal, twoside]{IEEEtran}  

\IEEEoverridecommandlockouts                              

\usepackage{graphicx}
\usepackage{amsfonts}
\usepackage{booktabs}
\usepackage{multirow}
\usepackage{pifont}
\usepackage{xcolor}
\usepackage{amsmath}
\usepackage{makecell}
\usepackage{arydshln}
\makeatletter
\let\NAT@parse\undefined
\makeatother
\usepackage{hyperref}

\begin{document}

\title{MGS-SLAM: Monocular Sparse Tracking and Gaussian Mapping with Depth Smooth Regularization}

\author{Pengcheng Zhu, Yaoming Zhuang, Baoquan Chen, Li Li, Chengdong Wu, and Zhanlin Liu
\thanks{Manuscript received May, 9, 2024; Revised July, 6, 2024; Accepted August, 28, 2024. This paper was recommended for publication by Editor Sven Behnke upon evaluation of the Associate Editor and Reviewers' comments. This research was supported in part by the National Natural Science Foundation of China (62403108, 42301256, U20A20197 and 61973063), the Liaoning Provincial Natural Science Foundation Joint Fund (2023-MSBA-075), the Ministry of Industry and Information Technology Project (TC220H05X-04), the Scientific Research Foundation of Liaoning Provincial Education Department (LJKQR20222509), the Fundamental Research Funds for the Central Universities (N2426005). (Corresponding Author: Yaoming Zhuang) Pengcheng Zhu and Yaoming Zhuang contributed to the paper equally and should be regarded as co-first authors.}
\thanks{Pengcheng Zhu, Yaoming Zhuang, Baoquan Chen, Chengdong Wu are with the Faculty of Robot Science and Engineering, College of Information Science and Engineering, Northeastern University, Shenyang 110819, China (e-mail: 2201005@stu.neu.edu.cn; zhuangyaoming@mail.neu.edu.cn;  2202035@stu.neu.edu.cn; wuchengdong@mail.neu.edu.cn).}
\thanks{Li Li is with the JangHo School of Architecture, Northeastern University, Shenyang 110819, China (e-mail: lili1118@mail.neu.edu.cn).}%
\thanks{Zhanlin Liu is with the AstrumU, Bellevue, Washington, 98004, USA (e-mail: Kevin.liu@astrumu.com).}%
\thanks{Digital Object Identifier (DOI): see top of this page.}
}

\markboth{IEEE Robotics and Automation Letters. Preprint Version. Accepted August, 2024}
{Zhu \MakeLowercase{\textit{et al.}}: MGS-SLAM: Monocular Sparse Tracking and Gaussian Mapping with Depth Smooth Regularization} 

\maketitle

\begin{abstract}
This letter introduces a novel framework for dense Visual Simultaneous Localization and Mapping (VSLAM) based on Gaussian Splatting. Recently, SLAM based on Gaussian Splatting has shown promising results. However, in monocular scenarios, the Gaussian maps reconstructed lack geometric accuracy and exhibit weaker tracking capability. To address these limitations, we jointly optimize sparse visual odometry tracking and 3D Gaussian Splatting scene representation for the first time. We obtain depth maps on visual odometry keyframe windows using a fast Multi-View Stereo (MVS) network for the geometric supervision of Gaussian maps. Furthermore, we propose a depth smooth loss and Sparse-Dense Adjustment Ring (SDAR) to reduce the negative effect of estimated depth maps and preserve the consistency in scale between the visual odometry and Gaussian maps. We have evaluated our system across various synthetic and real-world datasets. The accuracy of our pose estimation surpasses existing methods and achieves state-of-the-art. Additionally, it outperforms previous monocular methods in terms of novel view synthesis and geometric reconstruction fidelities.

\end{abstract}


\section{Introduction}

\IEEEPARstart{S}{imultaneous} Localization and Mapping (SLAM) is a key technology in robotics and autonomous driving. It aims to solve the problem of how robots determine their location and reconstruct maps of the environment in unknown scenes. The development of SLAM technology has gone through multiple stages, starting with the initial filter-based method \cite{monoslam}, advancing to graph optimization-based method \cite{orb-slam1}, and more recently, integrating deep learning. This integration has significantly improved the accuracy and robustness of SLAM systems. With the rapid development of deep learning technology, a new approach to SLAM technology has emerged, utilizing differentiable rendering. The initial applications of differentiable rendering-based SLAM utilized Neural Radiance Fields (NeRF) as their foundational construction method. NeRF, as detailed in \cite{NeRF}, employs neural networks to represent 3D scenes, enabling the synthesis of high-quality images and the recovery of dense geometric structures from multiple views. NeRF-based SLAM systems preserve detailed scene information during mapping, which enhances support for subsequent navigation and path planning. However, NeRF's approach requires multiple forward predictions for each pixel during image rendering, leading to significant computational redundancy. Consequently, this inefficiency prevents NeRF-based SLAM from operating in real-time, thus limiting its practicality for immediate downstream tasks.

\begin{figure}[t]
\centering
\setlength\abovecaptionskip{0pt}
\includegraphics[width=0.97\columnwidth]{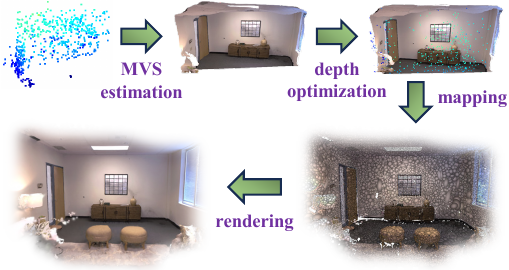}
\caption{Map reconstruction process by the proposed system. The prior depth map is estimated from the keyframes of sparse visual odometry and optimized by a sparse point cloud map, and the optimized depth map is used to construct a dense Gaussian map.}
\label{figure1}
\vspace{-12pt}
\end{figure}

Recently, a novel scene representation framework called 3D Gaussian Splatting \cite{3DGS} has demonstrated superior performance compared to NeRF. It features a more concise scene representation method and real-time rendering capability. This method not only delivers an accurate description of the scene but also offers a differentiable approach for optimizing the scene and camera poses. This opens up a new research direction for differentiable rendering-based SLAM. However, current Gaussian Splatting-based SLAM systems rely on the depth maps input to achieve precise geometric reconstruction, which constrains the scope of their application.

\begin{figure*}[t]
\centering
\setlength\abovecaptionskip{0pt}
\includegraphics[width=1.0\textwidth]{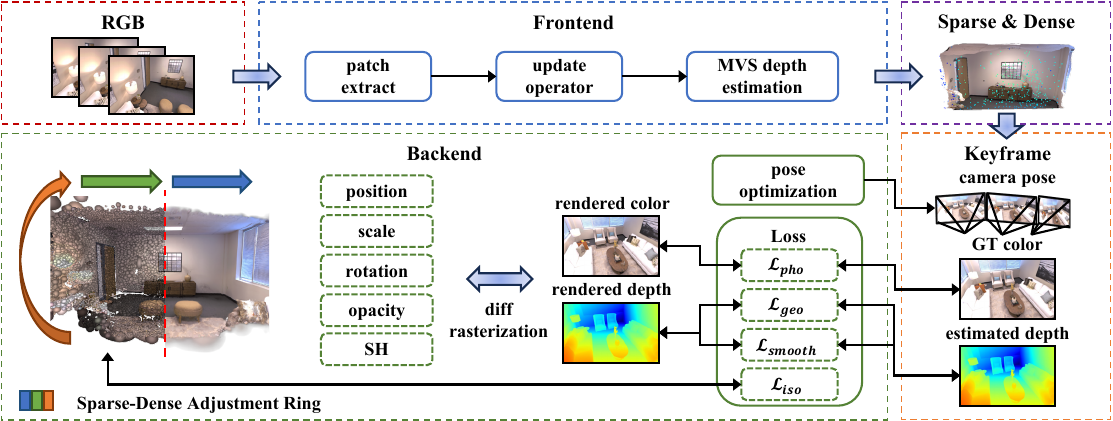}
\caption{\textbf{System pipeline.} The system inputs an RGB stream and operates frontend and backend processes in parallel. In the frontend, sparse visual odometry extracts patch features from images to estimate poses. These estimated poses and images are inputs to a pre-trained Multi-View Stereo (MVS) network, which estimates priori depth maps. In the backend, the estimated priori depth maps and images, coupled with poses from the frontend, are utilized as supervisory information to construct a Gaussian map. The frontend and backend maintain scale consistency through the SDAR strategy.}
\label{figure2}
\vspace{-10pt}
\end{figure*}

This letter presents MGS-SLAM, a novel monocular Gaussian Splatting-based SLAM system. This work introduces several groundbreaking advancements in the field of SLAM, which include integrating Gaussian Splatting techniques with sparse visual odometry, employing a pre-trained Multi-View Stereo (MVS) depth estimation network, pioneering a geometric smooth depth loss, and developing the SDAR strategy to ensure scale consistency. Together, these innovations significantly improve the accuracy and functionality of SLAM systems that rely solely on RGB image input. Fig. \ref{figure1} illustrates the map construction process: initially, sparse visual odometry constructs the sparse maps; subsequently, the MVS depth estimation network generates priori depth maps; these depth maps, along with the sparse point maps, are then refined through depth optimization in the SDAR; and finally, the Gaussian map is constructed using the optimized depth maps and depth smooth regularization loss.

The key contributions of the proposed system are summarized as follows:

\begin{itemize}
\item Introducing the first SLAM system that jointly optimizes sparse visual odometry poses and 3D Gaussian Splatting to achieve the accurate geometric reconstruction of Gaussian maps and pose tracking.
\item Developing a pre-trained Multi-View Stereo (MVS) depth estimation network that utilizes sparse odometry keyframes and their poses to estimate prior depth maps, thus providing crucial geometric constraints for Gaussian map reconstruction with only RGB image input.
\item Proposing a geometric depth smooth loss method to minimize the adverse impacts of inaccuracies in estimated prior depth maps on the Gaussian map and guide its alignment to correct geometric positions.
\item Proposing a Sparse-Dense Adjustment Ring (SDAR) strategy to unify the scale consistency of sparse visual odometry and dense Gaussian map.
\end{itemize}

\section{Related Works}

\textbf{Monocular Dense SLAM.} Over the past few decades, monocular dense SLAM technology has seen significant advancements. DTAM \cite{DTAM} pioneered one of the earliest real-time dense SLAM systems by performing parallel depth computations on GPU. To balance computational costs and accuracy, there are also semi-dense methods such as \cite{LSD-SLAM}, but these methods struggle to capture areas with poor texture. In the era of deep learning, DROID-SLAM \cite{DROID-SLAM} utilizes optical flow networks to establish dense pixel correspondences and achieve precise pose estimation. Another study \cite{TANDEM}, combines a real-time VO system with a Multi-View Stereo (MVS) network for parallel tracking and dense depth estimation, and then the Truncated Signed Distance Function (TSDF) is used to fuse depth maps and extract mesh. Codemapping \cite{codemapping} and Rosinol et al. \cite{probabilistic} incorporate sparse point cloud correction and volumetric fusion strategy on the estimated depth map to mitigate the impact of errors in the estimated depth map. We have also adopted a strategy for correcting the estimated depth map, but the difference is that we use a linear variance correction as depth optimization, which is less computations.

\textbf{Differentiable Rendering SLAM.} With the emergence of Neural Radiance Fields (NeRF) in 2020, numerous NeRF-based SLAM works have been proposed. iMAP \cite{imap} represented the pioneering work in NeRF-based SLAM, utilizing a dual-threading mode to track camera poses and execute mapping simultaneously. NICE-SLAM \cite{nice-slam} introduced feature grids based on iMAP, enabling NeRF-based SLAM to represent larger scenes. Subsequent works such as GO-SLAM \cite{go-slam} and Loopy-SLAM \cite{loopy-slam} incorporated global bundle adjustment (BA) and loop closure correction, further enhancing pose estimation accuracy and mapping performance. PLGSLAM \cite{plg-slam} proposes a progressive scene representation method to improve reconstruction and localization accuracy in large scenarios. Recently, 3D Gaussian Splatting has shown superior performance in 3D scene representation. It has fast rendering capability and is more suitable for online systems like SLAM. SplaTAM \cite{splatam} and GS-SLAM \cite{gs-slam} combine 3D Gaussian Splatting with SLAM, leveraging the realistic scene reconstruction ability of 3D Gaussian Splatting to surpass NeRF-based SLAM methods in rendering quality. Compact-SLAM \cite{compact-slam} proposes a compact 3D Gaussian Splatting SLAM system that reduces the number and the parameter size of Gaussian ellipsoids. NGM-SLAM \cite{ngm-slam} utilizes neural radiance field submaps for progressive scene expression, achieving effective loop closure detection. MonoGS \cite{MonoGS} and Photo-SLAM \cite{photo-slam} achieve monocular map reconstruction of Gaussian Splatting-based SLAM. However, existing Gaussian Splatting-based SLAM implementations typically require depth map input from RGB-D sensors to obtain accurate geometry reconstruction.

\begin{figure}[t]
\centering
\setlength\abovecaptionskip{0pt}
\includegraphics[width=1.0\columnwidth]{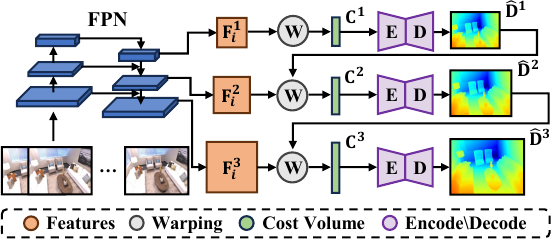}
\caption{The fast Multi-View Stereo network. The inputs of the network are images with poses from sparse visual odometry, image features are extracted by Feature Pyramid Network (FPN) and warped to the 2D cost volume. Finally, encoded and decoded to depth maps using coarse-to-fine strategy.}
\label{figure3}
\vspace{-12pt}
\end{figure}

\section{Methods}

Our approach utilizes RGB image as input, parallelly performing camera pose estimation and photorealistic dense mapping. As depicted in Fig. \ref{figure2}, the core idea of the approach is to use a pre-trained Multi-View Stereo (MVS) network to couple sparse VO and dense Gaussian Splatting mapping. Specifically, in the frontend part, tracking RGB image provides the backend with coarse camera poses and priori depth maps (Sec. \ref{sec3-A}). In the backend part, we represent the dense map using 3D Gaussian Splatting, and jointly optimize the dense map and the coarse poses from the frontend (Sec. \ref{sec3-B}). In the system components part, system initialization, selecting the keyframes for the system and correcting the scale between sparse point cloud map and dense Gaussian map by SDAR strategy are reported (Sec. \ref{sec3-C}).

\subsection{Sparse Visual Odometry Frontend}

\label{sec3-A}

To achieve more accurate camera pose tracking and provide dense depth geometry before backend mapping, the frontend of our framework is built on the Deep Patch Visual Odometry (DPVO) \cite{DPVO} algorithm. DPVO is a learning-based sparse monocular visual odometry method. Given an input RGB stream, the scene is represented as a collection of camera poses $\textbf{T} \in SE(3)^N$ and a series of square image patches \textbf{P} extracted from the images. The reprojection of a square patch $k$ taken from frame $i$ in frame $j$ can be formulated as: 
\begin{equation}
\textbf{P}^{ij}_k \sim \textbf{K}\textbf{T}_j\textbf{T}^{-1}_i\textbf{K}^{-1}\textbf{P}^i_k \label{eq1}
\end{equation}
where \textbf{K} refers to camera intrinsic matrix, $\textbf{P}^i_k = [u,v,1,d]^T$ denotes patch $k$ in frame $i$, and $[u,v]$ denote the pixel coordinates in images, $d$ denotes the inverse depth.

The core of DPVO is an update operator that computes the hidden state for each edge $(k,i,j) \in \varepsilon$. It optimizes the reprojection errors on the patch graph to predict a 2D correction vector $\delta^{ij}_k \in \mathbb{R}^2$ and confidence weight $\psi^{ij}_k \in \mathbb{R}^2$. Bundle Adjustment (BA) is performed using optical flow correction as a constraint, with iterative updates and refinement of camera poses and patch depths achieved through the non-linear least squares method. The cost function for bundle adjustment is as follows:
\begin{equation}
\sum_{(k,i,j) \in \varepsilon}\Vert \textbf{K}\textbf{T}_j\textbf{T}^{-1}_i\textbf{K}^{-1}\textbf{P}^i_k - [\bar{\textbf{P}}^{ij}_k+\delta^{ij}_k] \Vert^2_{\psi^{ij}_k} \label{eq2}
\end{equation}
where $\Vert\cdot\Vert_\psi$ represents Mahalanobis distance, $\bar{\textbf{P}}$ denotes the centre of patch.

\textbf{Multi-view priori depth estimation.} The backend dense Gaussian mapping requires the geometric supervision of depth maps to obtain the accurate geometric positions of Gaussians. In order to make monocular SLAM have the ability of geometric supervision, unlike the previous method \cite{COLMAP-Free}, we use a pre-trained Multi-View Stereo (MVS) network to estimate priori depth maps on the keyframes window of DPVO, the network is shown in Fig. \ref{figure3}. This method utilizes the geometric consistency of the MVS network to achieve the supervision of the geometric positions of Gaussians through only monocular RGB image input. Furthermore, our MVS network consists entirely of 2D convolutions with a coarse-to-fine structure that progressively refines the estimated priori depth map to reduce the runtime of the MVS network. Tab. \ref{table3} and Tab. \ref{recon} show that this method achieves better rendering and reconstruction performance.

To be more specific, the frame currently tracked by the sparse visual odometry is used as the reference image $\textbf{I}^0$. Additionally, we employ the previous $\textbf{N}$ keyframes as a series of original images $\textbf{I}^{n\in{1,..., N}}$. These images and their corresponding camera poses, serve as inputs to the MVS network. Utilizing the Feature Pyramid Network (FPN) module, we extract three layers of image features $\textbf{F}^s_i$ for each image, with $s$ denoting the layer index and $i$ representing the image index. In each layer, the original image features dot the reference image features by a differentiable warping operation to obtain a cost volume with dimensions $\textbf{D}\times\textbf{H}^s\times\textbf{W}^s$, and the priori depth map of each layer is obtained by 2D convolutions encoding and decoding. The estimated depth map of the previous layer is upsampled as the reference depth map of the next layer. The final depth map is estimated after three layers to achieve the coarse-to-fine effect.

Our MVS depth estimation network is trained on the ScanNet dataset \cite{scannet}. We train with the AdamW optimizer for 100k steps with a weight decay of $10^{-4}$, and a learning rate of $10^{-4}$ 
for 70k steps, $10^{-5}$ until 80k, then dropped to $10^{-6}$ for remainder,  which takes approximately 84 hours on two 24GB RTX3090 GPUs. We use a scale-invariant loss function to accommodate the relative poses of the sparse visual odometers:
\begin{equation}
\mathcal{L}^s_{si}=\sqrt{\frac{1}{H^sW^s}\sum_{i,j}(g^s_{i,j})^2-\frac{\lambda}{(H^sW^s)^2}(\sum_{i,j}g^s_{i,j})^2} \label{eq3}
\end{equation}
where $g^s_{i,j}={\uparrow_{gt}\log\hat{D}^s_{i,j}-\log{D}^{gt}_{i,j}}$. $D^{gt}_{i,j}$ denotes a ground truth depth map, which is aligned to the size of predicted depth $D^{s}_{i,j}$ by an upsampling operation $\uparrow_{gt}$. $\lambda$ is a constant 0.85.

In addition, the multi-view loss and the normal loss are added to the loss function to maintain the geometric consistency of depth estimation. The multi-view loss average absolute error on log depth over all valid points: 
\begin{equation}
\mathcal{L}^s_{mv}=\frac{1}{NH^sW^s}\sum_{n,i,j}\left|\uparrow_{gt}\log\textbf{T}_{0 n}(\hat{D}^s_{i,j})-\log{D^{gt}_{n,i,j}}\right| \label{eq4}
\end{equation}
\begin{equation}
\mathcal{L}^s_{normal}=\frac{1}{2H^sW^s}\sum_{i,j}(1-\hat{N}^s_{i,j}\cdot N^s_{i,j}) \label{eq5}
\end{equation}
where $\textbf{T}_{0 n}$ denotes the transformation matrix from the reference image to the original image $n$. $\hat{N}^s_{i,j}$ and $N^s_{i,j}$ respectively denote the prediction normals and ground truth normals. The final MVS depth estimation network loss is as follows:
\begin{equation}
\mathcal{L}=\sum^l_{s=0}\frac{1}{2^s}(\lambda_{si}\mathcal{L}^s_{si}+\lambda_{mv}\mathcal{L}^s_{mv}+\lambda_{normal}\mathcal{L}^s_{normal}) \label{eq6}
\end{equation}
where $l$ is 2, and  we assign the loss weights $\lambda_{si}$, $\lambda_{mv}$ and $\lambda_{normal}$ to 1.0, 0.2 and 1.0 respectively.

\subsection{3D Gaussian Splatting Mapping Backend}

\label{sec3-B}

The main responsibility of the backend is to further optimize the coarse poses from the frontend and map a Gaussian scene. The key to this thread is differentiable rendering and depth smooth regularisation loss, computing the loss between the renderings and the ground truth, and adjusting the coarse poses and Gaussian map by backward gradient propagation.

\textbf{Differentiable Gaussian map representation.} We use 3D Gaussian Splatting as a dense representation of the scene. The influence of a single 3D Gaussian $p_i \in \mathbb{R}^3$ in 3D scene is as follows:
\begin{equation}
f(p_i)=\sigma(o_i)\cdot \exp(-\frac{1}{2}(p_i-\mu_i)^T\Sigma^{-1}(p_i-\mu_i)) \label{eq7}
\end{equation}
where $o_i \in \mathbb{R}$ denotes the opacity of the Gaussian, $\mu_i \in \mathbb{R}^3$ is the centre of the Gaussian, $\Sigma=RSS^TR^T \in \mathbb{R}^{3,3}$ is the covariance matrix computed with $S \in \mathbb{R}^3$ scaling and $R \in \mathbb{R}^{3,3}$ components. The expression for the projection of a 3D Gaussian onto the image plane is as follows:
\begin{equation}
\mu_I=\pi(\textbf{T}_{CW}\cdot\mu_W)
\label{eq8}
\end{equation}
\begin{equation}
\Sigma_I=JW\Sigma_{W}W^TJ^T
\label{eq9}
\end{equation}
where $\pi(\cdot)$ denotes the projection of the 3D Gaussian center, $\textbf{T}_{CW} \in SE(3)$ is the the transformation matrix from world coordinate to camera coordinate in 3D space, $J$ is a linear approximation to the Jacobian matrix of the projective transformation, $W$ is the rotational component of $\textbf{T}_{CW}$. The Eq. (\ref{eq8}) and Eq. (\ref{eq9}) are differentiable, which ensures that the Gaussian map can be used with first-order gradient descent to continuously optimize the geometric and photometric of the map, allowing the map to be rendered as photo-realistic images. A single pixel color $C_p$ is rendered from $N$ Gaussians by splatting and blending: 
\begin{equation}
C_p=\sum_{i \in N}c_io_i\prod^{i-1}_{j=1}(1-o_j) \label{eq10}
\end{equation}
where $c_i$ is the color of Gaussian $i$, and $o_i$ is the opacity of Gaussian $i$.

\begin{figure}[t]
\centering
\setlength\abovecaptionskip{0pt}
\includegraphics[width=1.0\columnwidth]{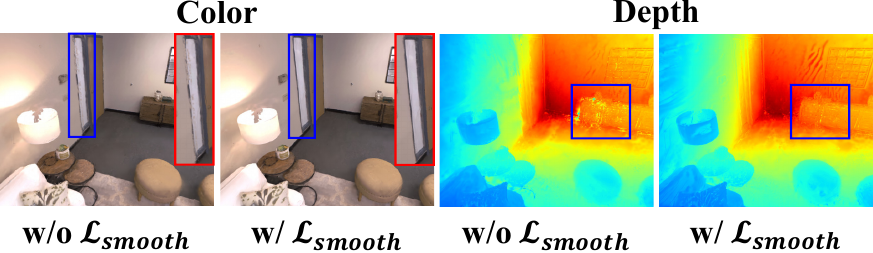}
\caption{Depth smooth regularization loss. Comparing the effect of having no depth smooth loss, there is better photometry and geometry with depth smooth loss, and bad photometry and geometry without depth smooth loss.}
\label{figure4}
\vspace{-12pt}
\end{figure}

\textbf{Mapping Optimization Losses.} We changed the loss function of the vanilla 3D Gaussian splatting and added more geometric constraints to make it more suitable for online mapping systems like SLAM. Specifically, our loss function consists of four components: photometric loss, depth geometric loss, depth smooth regularization loss and isotropic loss. In the photometric loss, the L1 loss is calculated between the rendered color image and the ground truth color image in the current camera pose $\textbf{T}_{CW}$: 
\begin{equation}
\mathcal{L}_{pho}=\Vert I(\mathcal{G}, \textbf{T}_{CW})-I_{gt} \Vert_1 \label{eq11}
\end{equation}
where $I(\mathcal{G}, \textbf{T}_{CW})$ is the rendered color image from Gaussians $\mathcal{G}$, and $I_{gt}$ is ground truth color image.

To improve the geometric accuracy of the Gaussian map, similar to Eq. (\ref{eq10}), We also rendered the depth:
\begin{equation}
D_p=\sum_{i \in N}z_io_i\prod^{i-1}_{j=1}(1-o_j) \label{eq12}
\end{equation}
where $z_i$ is the distance along the camera ray to the center $\mu_W$ of Gaussian $i$. Therefore, the depth geometric loss is as follows:
\begin{equation}
\mathcal{L}_{geo}=\Vert  D(\mathcal{G}, \textbf{T}_{CW})-\bar{D}_{d} \Vert_1 \label{eq13}
\end{equation}
where $D(\mathcal{G}, \textbf{T}_{CW})$ is the rendered depth map from Gaussians $\mathcal{G}$, $\bar{D}_{d}$ is the optimized priori depth map by SDAR strategy. The optimization process is in Sec. \ref{sec3-C}.

The prior depth maps obtained from the MVS network may not be entirely accurate. As depicted in Fig. \ref{figure4}, direct utilization of these depth maps leads to erroneous guidance in the geometric reconstruction of the Gaussian map. Similar to NeSLAM \cite{neslam}, we introduce the depth smooth regularization loss to reduce this erroneous guidance:
\begin{equation}
\mathcal{L}_{smooth}=\Vert d_{i,j-1}-d_{i,j} \Vert_2 + \Vert d_{i+1,j}-d_{i,j} \Vert_2 \label{eq14}
\end{equation}
where $d_{i,j}$ denotes the depth value of the pixel coordinate at $(i,j)$ in the rendering depth map. However, NeSLAM is an RGB-D SLAM system, which optimizes the noisy depth from RGB-D sensors by the denoising network and constraining the standard variance of depth to obtain better depth input. In contrast, we regularize the adjacent pixels between depth maps rendered from the Gaussian map, enabling the Gaussians to have better geometric positions.

The vanilla 3D Gaussian Splatting algorithm places no constraints on the Gaussians in the ray direction along the viewpoint. This has no effect on 3D reconstruction with fixed viewpoints. However, SLAM is an online mapping system, so this causes the Gaussians to elongate along the direction of the view ray, leading to the appearance of artifacts. To solve this problem, as well as \cite{MonoGS}, we also introduce isotropic loss:
\begin{equation}
\mathcal{L}_{iso}=\sum^{|\mathcal{G}|}_{i=1}\Vert s_i-\bar{s}_i\cdot1 \Vert_1 \label{eq15}
\end{equation}
where $s_i$ is the scaling of Gaussians, suppressing the elongation of the Gaussians by regularizing both the scaling and mean $\bar{s}_i$. The final mapping optimization loss function is as follows:
\begin{equation}
\mathcal{L}=\lambda_{p}\mathcal{L}_{pho}+\lambda_{g}\mathcal{L}_{geo}+\lambda_{s}\mathcal{L}_{smooth}+\lambda_{i}\mathcal{L}_{iso} \label{eq16}
\end{equation}
where we assign the loss weights $\lambda_{p}$, $\lambda_{g}$, $\lambda_{s}$ and $\lambda_{i}$ to 0.99, 0.01, 1.0 and 1.0 respectively.

\textbf{Camera poses optimization from the Gaussian map.} We use the camera poses $\textbf{T}^i_{CW}$ obtained from sparse visual odometry tracking in the frontend as the initial poses for Gaussian mapping in the backend. As in Eq. (\ref{eq10}) and Eq. (\ref{eq12}), we render the color image and depth map from the Gaussian map at the viewpoint of the current initial poses and compute the loss of renderings and the ground truth. Since this process is differentiable, the loss gradient is propagated to both the Gaussian map and the initial poses during the gradient backward process. The equation of the initial poses optimization update is as follows:
\begin{equation}
 \underset {\textbf{T}^i_{CW}, \mathcal{G}}{\mathrm {arg\,min}} \, \sum_{i=1}^{n} \mathcal{L}_{mapping} (\mathcal{G}, \textbf{T}^i_{CW}, I^i_{gt}, \bar{D}^i_{d}) \label{eq17}
\end{equation}
where $\mathcal{L}_{mapping}$ is the Eq. (\ref{eq16}), $I^i_{gt}$ and $\bar{D}^i_{d}$ are $i$th ground truth color image and optimized priori depth map from the viewpoint of $\textbf{T}^i_{CW}$ in mapping window. $n$ is mapping window size. Minimize the mapping loss to optimize both Gaussians $\mathcal{G}$ and initial poses $\textbf{T}^i_{CW}$ simultaneously.

\subsection{System Components}

\label{sec3-C}

\textbf{System initialization.} Similar to DPVO, The system uses 8 frames for initialization. The pose of the new frame is initialized using a constant velocity motion model. We add new patches and frames until 8 frames have been accumulated, and then run 12 iterations of the update operator. The 8 frames in the initialization are used as MVS network inputs to estimate the priori depth of the first frame. The backend uses the first priori depth as the foundation to initialize the Gaussian map.

\begin{figure}[t]
\centering
\setlength\abovecaptionskip{0pt}
\includegraphics[width=1.0\columnwidth]{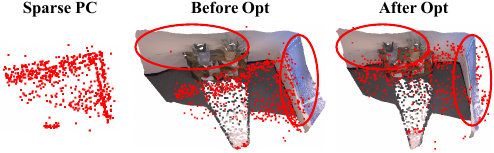}
\caption{Priori depth optimization. this optimization strategy in the SDAR is to correct the geometry of the priori depth map from the MVS network and align the scale with the sparse point cloud map.}
\label{figure5}
\vspace{-12pt}
\end{figure}

\textbf{Keyframe selection.} In the frontend tracking process, we always consider the 3 most recent frames as keyframes to fulfill the constant velocity motion model requirement. However, these 3 frames are not utilized for Gaussian mapping. Instead, we assess whether 4th frame satisfies Gaussian co-visibility criteria. If it does, we add it to the mapping process in the backend; otherwise, we discard this frame. This method can determine whether the tracked frame has new information exceeding a threshold, improve the efficiency of keyframe usage, and reduce memory consumption. Between two keyframes $i$, $j$, we define the co-visibility using Intersection of Union (IOU):
\begin{equation}
IOU_{cov}(i, j)=\frac{\left| \mathcal{G}_i \cap \mathcal{G}_j \right|}{\left| \mathcal{G}_i \cup \mathcal{G}_j \right|} \label{eq18}
\end{equation}
where $\mathcal{G}_i$, $\mathcal{G}_j$ are visible Gaussians in the viewpoints of frame $i$ and frame $j$. If IOU is less than a threshold, the system will create a new keyframe.

\textbf{Sparse-Dense Adjustment Ring.} We propose the Sparse-Dense Adjustment Ring (SDAR) strategy to achieve scale unification of the system. The method consists of three parts is as follows:

Firstly, We use a sparse point cloud map with better geometric accuracy to correct the priori depth map from the MVS network estimate. The priori depth map and the sparse depth map conform to the normal distribution of $\hat{D}_d\sim\mathcal{N}(\mu_d,\sigma^2_d)$ and $D_s\sim\mathcal{N}(\mu_s,\sigma^2_s)$. Align the priori depth map with the sparse depth map using the following equation:
\begin{equation}
\bar{D}_d=\frac{\sigma_s}{\hat{\sigma}_d}\hat{D}_d+\mu_d(\frac{\mu_s}{\hat{\mu}_d}-\frac{\sigma_s}{\hat{\sigma}_d}) \label{eq19}
\end{equation}
where $\hat{\mu}_d$ and $\hat{\sigma}_d$ are the mean and standard deviation statustics of the sparsified priori depth map extracted from $\hat{D}_d$ at the pixel coordinates of $D_s$. This strategy corrects the prior depth errors,  as shown in Fig. \ref{figure5}.

\begin{figure*}[t]
\centering
\setlength\abovecaptionskip{0pt}
\includegraphics[width=1.0\textwidth]{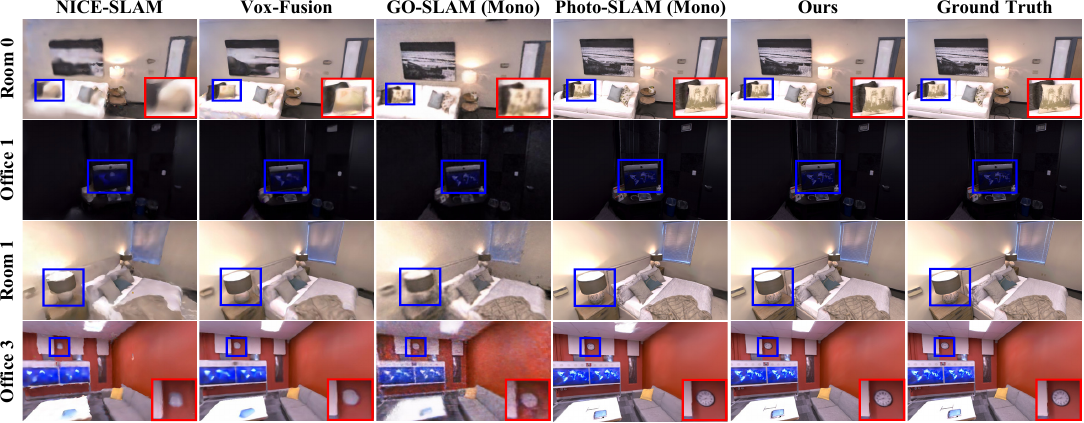}
\caption{The results of novel view rendering demonstrate the visualization outcomes on the Replica dataset for the proposed MGS-SLAM and other methods. Our system consistently generates significantly higher-quality and more realistic images than other monocular and RGB-D methods. This observation is further supported by quantitative results in Tab. \ref{table3}.}
\label{figure6}
\vspace{-5pt}
\end{figure*}

Secondly, we backproject the optimized prior depth map with RGB color into space, generating a new point cloud. Subsequently, downsampling is performed on this new point cloud. New Gaussians are then initialized with the downsampled point cloud and added to the Gaussian map.

Finally, to achieve scale closure, we leverage the real-time rendering capability of the Gaussian map to generate the depth map of the frame being tracked at the frontend. We then initialize the depth of the tracking frame's point cloud using this depth map. This strategy ensures that the frontend track aligns with the scale of the backend Gaussian map.

\section{Experiments}

We evaluate our proposed system on a series of real and synthetic datasets, including the TUM dataset \cite{TUM-RGBD}, Replica dataset \cite{replica} and ICL-NUIM dataset \cite{icl-nuim}. We compare the pose estimation accuracy (ATE), novel view rendering quality and geometric reconstruction quality with previous works, utilizing experimental results from papers or open-source code of these works. The experimental data from the source code represents the average of three runs. Additionally, we conduct some ablation studies to demonstrate the effectiveness of our system's components. Finally, we analyze the system runtime and memory.

\begin{table}[t]
\setlength{\abovecaptionskip}{0pt}
\caption{ATE [cm] results on TUM Dataset.}
\label{table1}
\centering
{\large
\resizebox{\linewidth}{!}{
\begin{tabular}{cccccccc}
\toprule
Input & Method & fr1/desk & fr1/desk2 & fr1/plant & fr2/xyz & fr3/office & Avg.\\
\midrule
\multirow{3}*{\rotatebox[origin=c]{90}{RGB-D}}
& SplaTAM & 3.35 & 6.54 & \underline{2.74} & \underline{1.24} & 5.16 & 3.81\\
& Co-SLAM & \underline{2.70} & \underline{4.31} & 4.74 & 1.90 & \underline{2.60} & \underline{3.25}\\
& ESLAM & \textbf{2.30} & \textbf{3.78} & \textbf{2.11} & \textbf{1.10} & \textbf{2.40} & \textbf{2.34}\\
\midrule
\multirow{6}*{\rotatebox[origin=c]{90}{Mono.}}
& DSO & 22.40 & 91.60 & 12.10 & 1.10 & 9.50 & 27.34\\
& DROID-VO & 5.20 & 9.90 & \textbf{2.80} & 10.70 & 7.30 & 7.18\\
& MonoGS & 4.15 & 7.16 & 7.82 & 4.79 & 4.39 & 5.66\\
& Photo-SLAM & \textbf{1.54} & 21.00 & 3.67 & 0.98 & \textbf{1.26} & 5.69\\
& DPVO & 3.80 & \underline{6.40} & 4.70 & \underline{0.54} & 7.00 & \underline{4.49}\\
& Ours & \underline{2.33} & \textbf{5.32} & \underline{3.55} & \textbf{0.44} & \underline{3.00} & \textbf{2.93}\\
\bottomrule
\end{tabular}
}
}
\end{table}

\begin{table}[t]
\setlength{\abovecaptionskip}{0pt}
\caption{ATE [cm] results on Replica Dataset}
\label{table2}
\centering
{\large
\resizebox{\linewidth}{!}{
\begin{tabular}{ccccccccccc}
\toprule
Input & Method & R0 & R1 & R2 & O0 & O1 & O2 & O3 & O4 & Avg.\\
\midrule
\multirow{5}*{\rotatebox[origin=c]{90}{RGB-D}}
& Co-SLAM & 0.70 & 0.95 & 1.35 & 0.59 & 0.55 & 2.03 & 1.56 & 0.72 & 1.06\\
& ESLAM & 0.71 & 0.70 & 0.52 & 0.57 & 0.55 & 0.58 & 0.72 & \underline{0.63} & 0.62\\
& NeSLAM & 0.60 & 0.93 & 0.52 & \textbf{0.41} & 0.43 & \underline{0.57} & 0.96 & 0.83 & 0.66\\
& GS-SLAM & \underline{0.48} & \underline{0.53} & \underline{0.33} & 0.52 & \underline{0.41} & 0.59 & \underline{0.46} & 0.70 & \underline{0.50}\\
& SplaTAM & \textbf{0.31} & \textbf{0.40} & \textbf{0.29} & \underline{0.47} & \textbf{0.27} & \textbf{0.29} & \textbf{0.32} & \textbf{0.55} & \textbf{0.36}\\
\midrule
\multirow{6}*{\rotatebox[origin=c]{90}{Mono.}}
& DROID-VO & 0.50 & 0.70 & \underline{0.30} & 0.98 & \underline{0.29} & 0.84 & \underline{0.45} & 1.53 & 0.70\\
& NICER-SLAM & 1.36 & 1.60 & 1.14 & 2.12 & 3.23 & 2.12 & 1.42 & 2.01 & 1.88\\
& MonoGS & 9.94 & 66.22 & 43.94 & 62.09 & 19.09 & 45.60 & 11.58 & 58.75 & 39.65\\
& Photo-SLAM & \textbf{0.35} & 1.18 & \textbf{0.23} & \underline{0.58} & 0.32 & 5.03 & 0.47 & 0.58 & 1.09\\
& DPVO & 0.49 & \underline{0.54} & 0.54 & 0.77 & 0.36 & \underline{0.57} & 0.46 & \underline{0.57} & \underline{0.54}\\
& Ours & \underline{0.36} & \textbf{0.35} & 0.32 & \textbf{0.35} & \textbf{0.28} & \textbf{0.26} & \textbf{0.32} & \textbf{0.34} & \textbf{0.32}\\
\bottomrule
\end{tabular}
}
}
\vspace{-10pt}
\end{table}

\begin{table}[t]
\setlength{\abovecaptionskip}{0pt}
\caption{ATE [cm] results on ICL-NUIM Dataset}
\label{icl}
\centering
{\large
\resizebox{\linewidth}{!}{
\begin{tabular}{ccccccccccc}
\toprule
Input & Method & L0 & L1 & L2 & L3 & O0 & O1 & O2 & O3  & Avg.\\
\midrule
\multirow{3}*{\rotatebox[origin=c]{90}{RGB-D}}
& Co-SLAM & 1.15 & 0.85 & \textbf{1.03} & 16.46 & 52.46 & 3.60 & 1.76 & 39.15 & 14.56\\
& ESLAM & \textbf{0.45} & \textbf{0.49} & 1.61 & \underline{5.84} & \textbf{0.42} & \underline{1.37} & \underline{1.01} & \textbf{0.46} & \underline{1.46}\\
& SplaTAM & \underline{0.53} & \underline{0.70} & \underline{1.13} & \textbf{4.63} & \textbf{0.42} & \textbf{1.03} & \textbf{0.92} & \underline{1.16} & \textbf{1.32}\\
\midrule
\multirow{6}*{\rotatebox[origin=c]{90}{Mono.}}
& DSO & 1.00 & 2.00 & 6.00 & 3.00 & 21.00 & 83.00 & 36.00 & 64.00 & 27.00\\
& DROID-VO & 1.00 & 12.30 & 7.20 & 3.20 & 9.50 & 4.10 & 84.20 & 50.40 & 21.49\\
& MonoGS & 6.40 & 21.21 & 31.40 & 100.76 & 13.87 & 35.76 & 24.73 & 73.42 & 38.44\\
& Photo-SLAM & \textbf{0.54} & 4.52 & \textbf{0.72} & \underline{0.98} & \underline{3.41} & 18.19 & \underline{1.54} & \underline{4.71} & \underline{4.33}\\
& DPVO & 0.60 & \underline{0.60} & 2.30 & 1.00 & 6.70 & \underline{1.20} & 1.70 & 63.50 & 9.70\\
& Ours & \underline{0.58} & \textbf{0.50} & \underline{1.82} & \textbf{0.77} & \textbf{1.46} & \textbf{1.01} & \textbf{1.19} & \textbf{1.49} & \textbf{1.10}\\
\bottomrule
\end{tabular}
}
}
\vspace{-10pt}
\end{table}

\subsection{Implementation Details}

We evaluate our proposed system and other methods on a desktop with an Intel Core i7 12700 processor running at 3.60GHz and a single NVIDIA GeForce RTX 3090. The size of input images is consistent with the dataset size in our system. Similar to Gaussian Splatting, mapping rasterization and gradient computations are implemented using CUDA. The remainder of our system pipeline is developed with PyTorch. For map optimization, we set the maximum gradient threshold to 0.0002 and the minimum opacity threshold to 0.65 for the Gaussians in the densify and prune operation.

\subsection{Camera Tracking Accuracy}

For camera tracking accuracy, we report the Root
Mean Square Error (RMSE) of the keyframe's Absolute Trajectory Error (ATE). We benchmark our system against other approaches. The comparative works are very comprehensive including traditional visual odometry DSO \cite{DSO}, learning-based visual odometry DROID-VO \cite{DROID-SLAM}, neural implicit-based NICER-SLAM \cite{nicer-slam}, NeSLAM \cite{neslam}, ESLAM \cite{eslam}, 
Co-SLAM \cite{co-slam} and more recently Gaussian Splatting-based SplaTAM \cite{splatam}, MonoGS \cite{MonoGS}, GS-SLAM \cite{gs-slam}, Photo-SLAM \cite{photo-slam}.

Tab. \ref{table1} shows the tracking results on the TUM dataset. The tracking accuracy of our system outperforms other monocular methods by 35\% and is comparable to ESLAM using RGB-D input. Tab. \ref{table2} and Tab. \ref{icl} show that our system achieved the best tracking performance compared to other systems including monocular and RGB-D. In addition, The experimental data from the tables show that our tracking performance is superior to the DPVO on which the frontend is based. This demonstrates the effectiveness of our combination of sparse visual odometry and Gaussian mapping in achieving a more robust and accurate SLAM system.

\begin{table}[t]
\setlength{\abovecaptionskip}{0pt}
\caption{Rendering performance on Replica Dataset. Best results are highlighted as \colorbox[rgb]{1.0,0.6,0.6}{first}, \colorbox[rgb]{1.0,0.8,0.6}{second}, and \colorbox[rgb]{1.0,0.98,0.6}{third}}
\label{table3}
\centering
{\large
\resizebox{\linewidth}{!}{
\begin{tabular}{cccccccccccc}
\toprule
Input & Method & Metric & R0 & R1 & R2 & O0 & O1 & O2 & O3 & O4 & Avg.\\
\midrule
\multirow{13.7}*{\rotatebox[origin=c]{90}{RGB-D}} 
& \multirow{3}*{\makecell[c]{NICE-\\SLAM}} & PSNR[dB]$\uparrow$ & 22.12 & \colorbox[rgb]{1.0,0.98,0.6}{22.47} & \colorbox[rgb]{1.0,0.98,0.6}{24.52} & \colorbox[rgb]{1.0,0.98,0.6}{29.07} & \colorbox[rgb]{1.0,0.8,0.6}{30.34} & 19.66 & 22.23 & 24.94 & \colorbox[rgb]{1.0,0.98,0.6}{24.42}\\
& & SSIM$\uparrow$ & \colorbox[rgb]{1.0,0.98,0.6}{0.689} & \colorbox[rgb]{1.0,0.98,0.6}{0.757} & \colorbox[rgb]{1.0,0.98,0.6}{0.814} & \colorbox[rgb]{1.0,0.98,0.6}{0.874} & \colorbox[rgb]{1.0,0.98,0.6}{0.886} & \colorbox[rgb]{1.0,0.98,0.6}{0.797} & 0.801 & \colorbox[rgb]{1.0,0.98,0.6}{0.856} & \colorbox[rgb]{1.0,0.98,0.6}{0.809}\\
& & LPIPS$\downarrow$ & 0.330 & \colorbox[rgb]{1.0,0.8,0.6}{0.271} & \colorbox[rgb]{1.0,0.6,0.6}{0.208} & \colorbox[rgb]{1.0,0.98,0.6}{0.229} & \colorbox[rgb]{1.0,0.6,0.6}{0.181} & \colorbox[rgb]{1.0,0.6,0.6}{0.235} & \colorbox[rgb]{1.0,0.8,0.6}{0.209} & \colorbox[rgb]{1.0,0.6,0.6}{0.198} & \colorbox[rgb]{1.0,0.6,0.6}{0.233}\\
\noalign{\vskip 2pt}\cdashline{2-12}\noalign{\vskip 2pt}
& \multirow{3}*{\makecell[c]{Vox-\\Fusion}} & PSNR[dB]$\uparrow$ & \colorbox[rgb]{1.0,0.98,0.6}{22.39} & 22.36 & 23.92 & 27.79 & 29.83 & \colorbox[rgb]{1.0,0.98,0.6}{20.33} & \colorbox[rgb]{1.0,0.98,0.6}{23.47} & \colorbox[rgb]{1.0,0.98,0.6}{25.21} & 24.41\\
& & SSIM$\uparrow$ & 0.683 & 0.751 & 0.798 & 0.857 & 0.876 & 0.794 & \colorbox[rgb]{1.0,0.98,0.6}{0.803} & 0.847 & 0.801\\
& & LPIPS$\downarrow$ & \colorbox[rgb]{1.0,0.6,0.6}{0.303} & \colorbox[rgb]{1.0,0.6,0.6}{0.269} & \colorbox[rgb]{1.0,0.8,0.6}{0.234} & 0.241 & \colorbox[rgb]{1.0,0.8,0.6}{0.184} & \colorbox[rgb]{1.0,0.98,0.6}{0.243} & \colorbox[rgb]{1.0,0.98,0.6}{0.213} & \colorbox[rgb]{1.0,0.8,0.6}{0.199} & \colorbox[rgb]{1.0,0.8,0.6}{0.236}\\
\noalign{\vskip 2pt}\cdashline{2-12}\noalign{\vskip 2pt}
& \multirow{3}*{\makecell[c]{ESLAM}} & PSNR[dB]$\uparrow$ & \colorbox[rgb]{1.0,0.8,0.6}{25.32} & \colorbox[rgb]{1.0,0.8,0.6}{27.77} & \colorbox[rgb]{1.0,0.6,0.6}{29.08} & \colorbox[rgb]{1.0,0.8,0.6}{33.71} & \colorbox[rgb]{1.0,0.98,0.6}{30.20} & \colorbox[rgb]{1.0,0.8,0.6}{28.09} & \colorbox[rgb]{1.0,0.6,0.6}{28.77} & \colorbox[rgb]{1.0,0.8,0.6}{29.71} & \colorbox[rgb]{1.0,0.8,0.6}{29.08}\\
& & SSIM$\uparrow$ & \colorbox[rgb]{1.0,0.8,0.6}{0.875} & \colorbox[rgb]{1.0,0.8,0.6}{0.902} & \colorbox[rgb]{1.0,0.6,0.6}{0.932} & \colorbox[rgb]{1.0,0.8,0.6}{0.960} & \colorbox[rgb]{1.0,0.8,0.6}{0.923} & \colorbox[rgb]{1.0,0.6,0.6}{0.943} & \colorbox[rgb]{1.0,0.6,0.6}{0.948} & \colorbox[rgb]{1.0,0.8,0.6}{0.945} & \colorbox[rgb]{1.0,0.8,0.6}{0.928}\\
& & LPIPS$\downarrow$ & \colorbox[rgb]{1.0,0.8,0.6}{0.313} & 0.298 & \colorbox[rgb]{1.0,0.98,0.6}{0.248} & \colorbox[rgb]{1.0,0.6,0.6}{0.184} & 0.228 & \colorbox[rgb]{1.0,0.8,0.6}{0.241} & \colorbox[rgb]{1.0,0.6,0.6}{0.196} & \colorbox[rgb]{1.0,0.98,0.6}{0.204} & \colorbox[rgb]{1.0,0.98,0.6}{0.239}\\
\noalign{\vskip 2pt}\cdashline{2-12}\noalign{\vskip 2pt}
& \multirow{3}*{\makecell[c]{Co-\\SLAM}} & PSNR[dB]$\uparrow$ & \colorbox[rgb]{1.0,0.6,0.6}{27.27} & \colorbox[rgb]{1.0,0.6,0.6}{28.45} & \colorbox[rgb]{1.0,0.8,0.6}{29.06} & \colorbox[rgb]{1.0,0.6,0.6}{34.14} & \colorbox[rgb]{1.0,0.6,0.6}{34.87} & \colorbox[rgb]{1.0,0.6,0.6}{28.43} & \colorbox[rgb]{1.0,0.8,0.6}{28.76} & \colorbox[rgb]{1.0,0.6,0.6}{30.91} & \colorbox[rgb]{1.0,0.6,0.6}{30.24}\\
& & SSIM$\uparrow$ & \colorbox[rgb]{1.0,0.6,0.6}{0.910} & \colorbox[rgb]{1.0,0.6,0.6}{0.909} & \colorbox[rgb]{1.0,0.6,0.6}{0.932} & \colorbox[rgb]{1.0,0.6,0.6}{0.961} & \colorbox[rgb]{1.0,0.6,0.6}{0.969} & \colorbox[rgb]{1.0,0.8,0.6}{0.938} & \colorbox[rgb]{1.0,0.8,0.6}{0.941} & \colorbox[rgb]{1.0,0.6,0.6}{0.955} & \colorbox[rgb]{1.0,0.6,0.6}{0.939}\\
& & LPIPS$\downarrow$ & \colorbox[rgb]{1.0,0.98,0.6}{0.324} & \colorbox[rgb]{1.0,0.98,0.6}{0.294} & 0.266 & \colorbox[rgb]{1.0,0.8,0.6}{0.209} & \colorbox[rgb]{1.0,0.98,0.6}{0.196} & 0.258 & 0.229 & 0.236 & 0.252\\
\midrule
\multirow{17}*{\rotatebox[origin=c]{90}{Mono.}}
& \multirow{3}*{\makecell[c]{GO-\\SLAM}} & PSNR[dB]$\uparrow$ & 23.25 & 20.70 & 21.08 & 21.44 & 22.59 & 22.33 & 22.19 & 22.76 & 22.04\\
& & SSIM$\uparrow$ & 0.712 & 0.739 & 0.708 & 0.761 & 0.726 & 0.740 & 0.752 & 0.722 & 0.733\\
& & LPIPS$\downarrow$ & \colorbox[rgb]{1.0,0.98,0.6}{0.222} & 0.492 & 0.317 & 0.319 & 0.269 & 0.434 & 0.396 & 0.385 & 0.354\\
\noalign{\vskip 2pt}\cdashline{2-12}\noalign{\vskip 2pt}
& \multirow{3}*{\makecell[c]{NICER-\\SLAM}} & PSNR[dB]$\uparrow$ & \colorbox[rgb]{1.0,0.98,0.6}{25.33} & 23.92 & \colorbox[rgb]{1.0,0.98,0.6}{26.12} & 28.54 & 25.86 & 21.95 & \colorbox[rgb]{1.0,0.98,0.6}{26.13} & \colorbox[rgb]{1.0,0.98,0.6}{25.47} & \colorbox[rgb]{1.0,0.98,0.6}{25.41}\\
& & SSIM$\uparrow$ & 0.751 & 0.771 & 0.831 & 0.866 & \colorbox[rgb]{1.0,0.98,0.6}{0.852} & 0.820 & \colorbox[rgb]{1.0,0.98,0.6}{0.856} & \colorbox[rgb]{1.0,0.98,0.6}{0.865} & 0.827\\
& & LPIPS$\downarrow$ & 0.250 & \colorbox[rgb]{1.0,0.98,0.6}{0.215} & \colorbox[rgb]{1.0,0.98,0.6}{0.176} & \colorbox[rgb]{1.0,0.98,0.6}{0.172} & \colorbox[rgb]{1.0,0.98,0.6}{0.178} & \colorbox[rgb]{1.0,0.98,0.6}{0.195} & \colorbox[rgb]{1.0,0.98,0.6}{0.162} & \colorbox[rgb]{1.0,0.98,0.6}{0.177} & \colorbox[rgb]{1.0,0.98,0.6}{0.191}\\
\noalign{\vskip 2pt}\cdashline{2-12}\noalign{\vskip 2pt}
& \multirow{3}*{\makecell[c]{Mono\\GS}} & PSNR[dB]$\uparrow$ & 25.11 & \colorbox[rgb]{1.0,0.98,0.6}{24.66} & 22.30 & \colorbox[rgb]{1.0,0.98,0.6}{28.76} & \colorbox[rgb]{1.0,0.98,0.6}{29.17} & \colorbox[rgb]{1.0,0.98,0.6}{23.74} & 23.66 & 23.99 & 25.17\\
& & SSIM$\uparrow$ & \colorbox[rgb]{1.0,0.98,0.6}{0.790} & \colorbox[rgb]{1.0,0.98,0.6}{0.790} & \colorbox[rgb]{1.0,0.98,0.6}{0.843} & \colorbox[rgb]{1.0,0.98,0.6}{0.884} & \colorbox[rgb]{1.0,0.98,0.6}{0.852} & \colorbox[rgb]{1.0,0.98,0.6}{0.840} & 0.855 & 0.863 & \colorbox[rgb]{1.0,0.98,0.6}{0.840}\\
& & LPIPS$\downarrow$ & 0.260 & 0.360 & 0.351 & 0.293 & 0.274 & 0.290 & 0.216 & 0.340 & 0.298\\
\noalign{\vskip 2pt}\cdashline{2-12}\noalign{\vskip 2pt}
& \multirow{3}*{\makecell[c]{Photo-\\SLAM}} & PSNR[dB]$\uparrow$ & \colorbox[rgb]{1.0,0.8,0.6}{29.07} & \colorbox[rgb]{1.0,0.8,0.6}{31.02} & \colorbox[rgb]{1.0,0.8,0.6}{31.22} & \colorbox[rgb]{1.0,0.8,0.6}{35.23} & \colorbox[rgb]{1.0,0.6,0.6}{35.11} & \colorbox[rgb]{1.0,0.8,0.6}{29.70} & \colorbox[rgb]{1.0,0.8,0.6}{31.20} & \colorbox[rgb]{1.0,0.8,0.6}{31.27} & \colorbox[rgb]{1.0,0.8,0.6}{31.73}\\
& & SSIM$\uparrow$ & \colorbox[rgb]{1.0,0.8,0.6}{0.845} & \colorbox[rgb]{1.0,0.6,0.6}{0.902} & \colorbox[rgb]{1.0,0.6,0.6}{0.923} & \colorbox[rgb]{1.0,0.6,0.6}{0.948} & \colorbox[rgb]{1.0,0.6,0.6}{0.942} & \colorbox[rgb]{1.0,0.6,0.6}{0.907} & \colorbox[rgb]{1.0,0.8,0.6}{0.915} & \colorbox[rgb]{1.0,0.8,0.6}{0.930} & \colorbox[rgb]{1.0,0.8,0.6}{0.914}\\
& & LPIPS$\downarrow$ & \colorbox[rgb]{1.0,0.8,0.6}{0.186} & \colorbox[rgb]{1.0,0.8,0.6}{0.125} & \colorbox[rgb]{1.0,0.8,0.6}{0.127} & \colorbox[rgb]{1.0,0.8,0.6}{0.109} & \colorbox[rgb]{1.0,0.8,0.6}{0.121} & \colorbox[rgb]{1.0,0.8,0.6}{0.173} & \colorbox[rgb]{1.0,0.8,0.6}{0.137} & \colorbox[rgb]{1.0,0.8,0.6}{0.120} & \colorbox[rgb]{1.0,0.8,0.6}{0.137}\\
\noalign{\vskip 2pt}\cdashline{2-12}\noalign{\vskip 2pt} 
& \multirow{3}*{Ours} & PSNR[dB]$\uparrow$ & \colorbox[rgb]{1.0,0.6,0.6}{29.91} & \colorbox[rgb]{1.0,0.6,0.6}{31.06} & \colorbox[rgb]{1.0,0.6,0.6}{31.49} & \colorbox[rgb]{1.0,0.6,0.6}{35.51} & \colorbox[rgb]{1.0,0.8,0.6}{34.25} & \colorbox[rgb]{1.0,0.6,0.6}{30.83} & \colorbox[rgb]{1.0,0.6,0.6}{31.86} & \colorbox[rgb]{1.0,0.6,0.6}{34.38} & \colorbox[rgb]{1.0,0.6,0.6}{32.41}\\
& & SSIM$\uparrow$ & \colorbox[rgb]{1.0,0.6,0.6}{0.894} & \colorbox[rgb]{1.0,0.8,0.6}{0.895} & \colorbox[rgb]{1.0,0.8,0.6}{0.913} & \colorbox[rgb]{1.0,0.8,0.6}{0.941} & \colorbox[rgb]{1.0,0.8,0.6}{0.930} & \colorbox[rgb]{1.0,0.8,0.6}{0.906} & \colorbox[rgb]{1.0,0.6,0.6}{0.919} & \colorbox[rgb]{1.0,0.6,0.6}{0.945} & \colorbox[rgb]{1.0,0.6,0.6}{0.918}\\
& & LPIPS$\downarrow$ & \colorbox[rgb]{1.0,0.6,0.6}{0.084} & \colorbox[rgb]{1.0,0.6,0.6}{0.086} & \colorbox[rgb]{1.0,0.6,0.6}{0.081} & \colorbox[rgb]{1.0,0.6,0.6}{0.070} & \colorbox[rgb]{1.0,0.6,0.6}{0.114} & \colorbox[rgb]{1.0,0.6,0.6}{0.120} & \colorbox[rgb]{1.0,0.6,0.6}{0.074} & \colorbox[rgb]{1.0,0.6,0.6}{0.077} & \colorbox[rgb]{1.0,0.6,0.6}{0.088}\\
\bottomrule
\end{tabular}
}
}
\end{table}

\begin{table}[t]
\setlength{\abovecaptionskip}{0pt}
\caption{Reconstruction performance on Replica Dataset. Best results are highlighted as \colorbox[rgb]{1.0,0.6,0.6}{first}, \colorbox[rgb]{1.0,0.8,0.6}{second}, and \colorbox[rgb]{1.0,0.98,0.6}{third}}
\label{recon}
\centering
\resizebox{\linewidth}{!}{
\begin{tabular}{cccccc}
\toprule
Input & Method & Depth L1[cm]$\downarrow$ & Acc.[cm]$\downarrow$ & Comp.[cm]$\downarrow$ & Comp. Ratio[$<$5cm]$\uparrow$ \\
\midrule
\multirow{6}*{\rotatebox[origin=c]{90}{Mono.}}
& MonoGS & 36.58 & 74.02 & 19.30 & 37.51\\
& Photo-SLAM & \colorbox[rgb]{1.0,0.98,0.6}{19.73} & 53.70 & 8.08 & 49.46\\
& GO-SLAM & \colorbox[rgb]{1.0,0.6,0.6}{4.39} & \colorbox[rgb]{1.0,0.8,0.6}{3.81} & \colorbox[rgb]{1.0,0.98,0.6}{4.79} & \colorbox[rgb]{1.0,0.98,0.6}{78.00}\\
& NICER-SLAM & - & \colorbox[rgb]{1.0,0.6,0.6}{3.65} & \colorbox[rgb]{1.0,0.8,0.6}{4.16} &\colorbox[rgb]{1.0,0.8,0.6}{79.37}\\
& Ours & \colorbox[rgb]{1.0,0.8,0.6}{7.77} & \colorbox[rgb]{1.0,0.98,0.6}{7.51} & \colorbox[rgb]{1.0,0.6,0.6}{3.64} & \colorbox[rgb]{1.0,0.6,0.6}{82.71}\\
\bottomrule
\end{tabular}
}
\vspace{-10pt}
\end{table}

\begin{table}[t]
\setlength{\abovecaptionskip}{0pt}
\caption{Mapping losses ablation on Office 0}
\label{table4}
\centering
\resizebox{\linewidth}{!}{
\begin{tabular}{cccccc}
\toprule
$\mathcal{L}_{geo}$ & $\mathcal{L}_{smooth}$ & $\mathcal{L}_{iso}$ & ATE[cm]$\downarrow$ & PSNR[dB]$\uparrow$ & Depth L1[cm]$\downarrow$\\
\midrule
\textcolor{red}{\ding{55}} & \textcolor{red}{\ding{55}} & \textcolor{red}{\ding{55}} & 0.53 & 31.21 & 25.46\\
\textcolor{green}{\ding{51}} & \textcolor{red}{\ding{55}} & \textcolor{red}{\ding{55}} & 0.45 & 33.80 & 11.09\\
\textcolor{green}{\ding{51}} & \textcolor{green}{\ding{51}} & \textcolor{red}{\ding{55}} & 0.40 & 33.88 & 7.21\\
\textcolor{green}{\ding{51}} & \textcolor{green}{\ding{51}} & \textcolor{green}{\ding{51}} & \textbf{0.35} & \textbf{34.85} & \textbf{5.37}\\
\bottomrule
\end{tabular}
}
\end{table}

\begin{table}[!t]
\setlength{\abovecaptionskip}{0pt}
\caption{Sparse-Dense Adjustment Ring ablation on Office 0}
\label{table5}
\centering
\resizebox{\linewidth}{!}{
\begin{tabular}{cccccc}
\toprule
Comp. 1 & Comp. 2 & Comp. 3 & ATE[cm]$\downarrow$ & PSNR[dB]$\uparrow$ & Depth L1[cm]$\downarrow$\\
\midrule
\textcolor{red}{\ding{55}} & \textcolor{red}{\ding{55}} & \textcolor{red}{\ding{55}} & 0.61 & 28.66 & 15.55\\
\textcolor{green}{\ding{51}} & \textcolor{red}{\ding{55}} & \textcolor{red}{\ding{55}} & 0.49 & 29.53 & 11.01\\
\textcolor{green}{\ding{51}} & \textcolor{green}{\ding{51}} & \textcolor{red}{\ding{55}} & 0.41 & 33.22 & 5.56\\
\textcolor{green}{\ding{51}} & \textcolor{green}{\ding{51}} & \textcolor{green}{\ding{51}} & \textbf{0.35} & \textbf{34.85} & \textbf{5.37}\\
\bottomrule
\end{tabular}
}
\end{table}

\subsection{Novel View Rendering}

We evaluated the methods for novel view rendering on Replica. To evaluate map quality, we report standard photometric rendering quality metrics (PSNR, SSIM and LPIPS). The methods we are comparing have RGB-D input and monocular input. NICE-SLAM \cite{nice-slam}, Vox-Fusion \cite{vox-fusion}, ESLAM \cite{eslam} and Co-SLAM \cite{co-slam} are neural implicit-based RGB-D input and the rest are monocular input. We take the average of frames other than keyframes to evaluate rendering quality. Tab. \ref{table3} shows the results, our proposed system performs state-of-the-art in most scenes. The visualization of the rendering is shown in Fig. \ref{figure6}, where the quality of our rendered image is higher than the other methods and almost indistinguishable from the ground truth.

\subsection{Geometric Reconstruction}

We evaluated the methods for geometric reconstruction on Replica. The methods evaluated are all monocular differentiable rendering SLAM approaches. We report standard mesh geometric reconstruction metrics (Depth L1, Accuracy, Completion, Completion Ratio). Tab. \ref{recon} shows the results, our method achieved the best results in terms of Completion and Completion Ratio metrics. It is worth noting that our geometric reconstruction performance is 50\% higher than other monocular 3D Gaussian Splatting-based SLAM, which proves the effectiveness of our method in utilizing the MVS network to promote geometric reconstruction. Furthermore, this better geometric reconstruction also improves the rendering.

\subsection{Ablative Analysis}

\textbf{Mapping losses ablation.} We changed the loss function of the vanilla 3D Gaussian Splatting by introducing depth loss, smooth loss, and isotropic loss. As shown in Tab. \ref{table4}, we did an ablation study of these losses. The results show that all these losses contribute to the accuracy improvement of the system. It is worth noting that the incorrect geometric guidance caused by the depth loss using the prior depth maps was corrected after adding depth smoothing loss.

\begin{table}[!t]
\setlength{\abovecaptionskip}{0pt}
\caption{Runtime and Memory analysis on TUM and Replica Datasets}
\label{table6}
\centering
\resizebox{\linewidth}{!}{
\begin{tabular}{cccccccc}
\toprule
Dataset & Method & Tra/It.$\downarrow$ & Map/It.$\downarrow$ & Tra/Fr.$\downarrow$ & Map/Fr.$\downarrow$ & Ren. FPS$\uparrow$ & Mem.$\downarrow$\\
\midrule
\multirow{4}*{\rotatebox[origin=c]{90}{TUM}}
& SplaTAM & \underline{14.28ms} & 16.77ms & 2.85s & \textbf{0.50s} & 526.32 & 42.31MB\\
& MonoGS & \textbf{6.78ms} & 12.67ms & 0.65s & 1.90s & 1126.10 & \underline{2.80MB}\\
& Photo-SLAM & - & \textbf{8.91ms} & \textbf{33.33ms} & - & \textbf{1648.20} & 12.77MB\\
& Ours & - & \underline{11.90ms} & \underline{35.17ms} & \underline{1.85s} & \underline{1173.21} & \textbf{1.96MB}\\
\midrule
\multirow{4}*{\rotatebox[origin=c]{90}{Replica}}
& SplaTAM & \underline{25.43ms} & 23.80ms & 2.25s & \textbf{1.43s} & 125.64 & 273.09MB\\
& MonoGS & \textbf{10.78ms} & 20.50ms & 1.10s & 3.07s & 769.00 & 24.50MB\\
& Photo-SLAM & - & \textbf{15.18ms} & \textbf{37.45ms} & - & \textbf{911.26} & \underline{22.21MB}\\
& Ours & - & \underline{18.98ms} & \underline{38.41ms} & \underline{2.97s} & \underline{776.50} & \textbf{20.90MB}\\
\bottomrule
\end{tabular}
}
\vspace{-10pt}
\end{table}

\begin{figure}[t]
\centering
\setlength\abovecaptionskip{0pt}
\includegraphics[width=0.95\columnwidth]{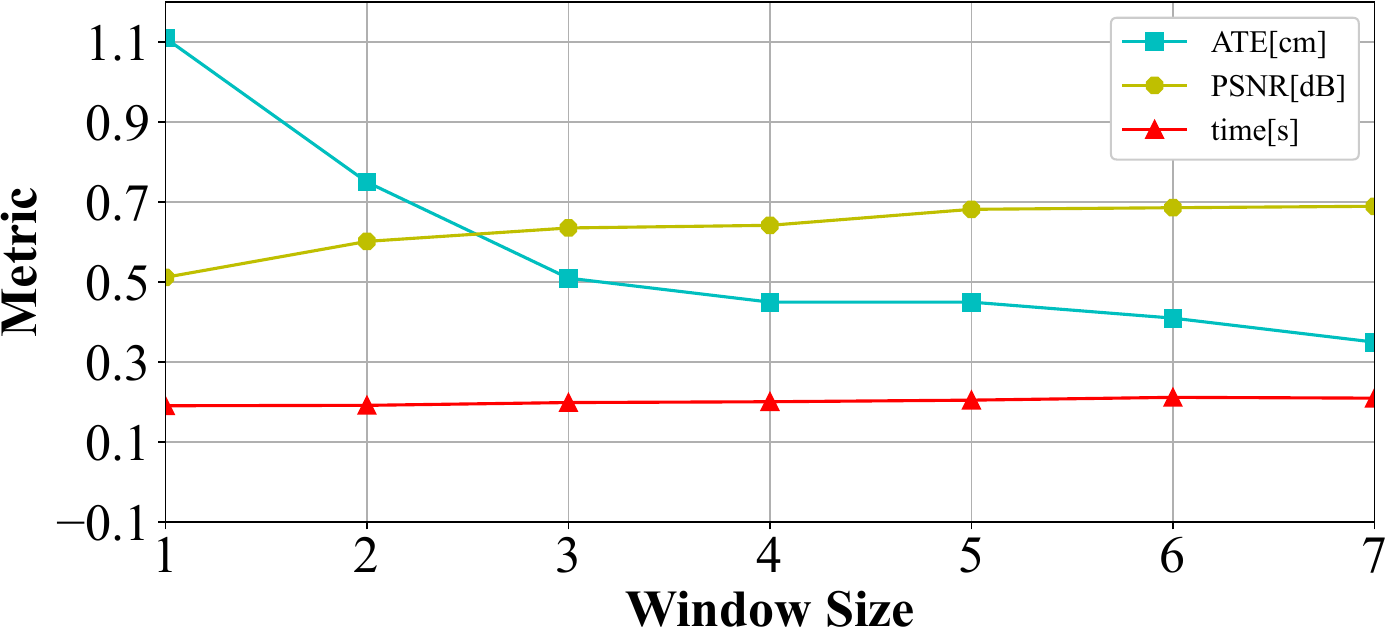}
\caption{MVS window analysis on Office 0. The MVS window size is a hyperparameter that allows for finding a balance between speed, tracking, and rendering quality. PSNR is divided by 50.}
\label{figure7}
\vspace{-12pt}
\end{figure}

\textbf{Sparse-Dense Adjustment Ring ablation.} We propose the Sparse-Dense Adjustment Ring (SDAR) strategy to unify the frontend and backend scales. This strategy comprises three components (Sec. \ref{sec3-C}). We conducted an ablation study of these three components to demonstrate their effect on the system. As shown in Tab. \ref{table5}, the contribution of SDAR to the system is mainly in the tracking accuracy ATE. The tracking accuracy of the system is similar to DPVO without the SDAR strategy. 

\textbf{MVS window analysis.} As depicted in Fig. \ref{figure7}, we have analyzed the effect of different window sizes of MVS on the accuracy and speed of the system. Since our MVS network consists of 2D convolutions, increasing the window size has little effect on inference time. However, increasing the window size improves the system's tracking accuracy and mapping quality. This is because more keyframes with different views provide additional geometrical cues.

\subsection{Runtime and Memory Analysis}

As shown in Tab. \ref{table6}, We quoted the method \cite{compact-slam} to analyze the runtime and memory of our system and compare it to other methods on the TUM and Replica datasets. The memory is the memory usage of the checkpoint. Some methods do not use this metric and are represented by shorter lines. The metric of tracking each frame contains the inference time of the MVS network in our method, and the MVS network runs on keyframes. The results show that our tracking speed is similar to Photo-SLAM. However, our method achieved better geometry at the expense of tracking time, resulting in more compact checkpoint and the best memory utilization.

\section{Conclusions}

This letter introduces MGS-SLAM, a novel Gaussian Splatting-based SLAM framework. For the first time, our framework jointly optimizes sparse visual odometry tracking and 3D Gaussian mapping, enhancing tracking accuracy and geometric reconstruction precision of Gaussian maps when only given RGB image input. We develop a lightweight MVS depth estimation network to facilitate this integration. Additionally, we propose the Sparse-Dense Adjustment Ring (SDAR) strategy to adjust the scale between the sparse map and the Gaussian map. Comparative evaluations demonstrate that our approach achieves state-of-the-art accuracy compared to previous methods. We believe that this innovative method will bring some inspiration to future works.

\addtolength{\textheight}{-12cm}   





\bibliographystyle{IEEEtran}
\bibliography{myrefs}

\begin{thebibliography}{10}
\providecommand{\url}[1]{#1}
\csname url@rmstyle\endcsname
\providecommand{\newblock}{\relax}
\providecommand{\bibinfo}[2]{#2}
\providecommand\BIBentrySTDinterwordspacing{\spaceskip=0pt\relax}
\providecommand\BIBentryALTinterwordstretchfactor{4}
\providecommand\BIBentryALTinterwordspacing{\spaceskip=\fontdimen2\font plus
\BIBentryALTinterwordstretchfactor\fontdimen3\font minus \fontdimen4\font\relax}
\providecommand\BIBforeignlanguage[2]{{%
\expandafter\ifx\csname l@#1\endcsname\relax
\typeout{** WARNING: IEEEtran.bst: No hyphenation pattern has been}%
\typeout{** loaded for the language `#1'. Using the pattern for}%
\typeout{** the default language instead.}%
\else
\language=\csname l@#1\endcsname
\fi
#2}}

\bibitem{monoslam}
A.~J. Davison, I.~D. Reid, N.~D. Molton, and O.~Stasse, ``Monoslam: Real-time single camera slam,'' \emph{IEEE transactions on pattern analysis and machine intelligence}, vol.~29, no.~6, pp. 1052--1067, 2007.

\bibitem{orb-slam1}
R.~Mur-Artal, J.~M.~M. Montiel, and J.~D. Tardos, ``Orb-slam: a versatile and accurate monocular slam system,'' \emph{IEEE transactions on robotics}, vol.~31, no.~5, pp. 1147--1163, 2015.

\bibitem{NeRF}
B.~Mildenhall, P.~P. Srinivasan, M.~Tancik, J.~T. Barron, R.~Ramamoorthi, and R.~Ng, ``Nerf: Representing scenes as neural radiance fields for view synthesis,'' \emph{Communications of the ACM}, vol.~65, no.~1, pp. 99--106, 2021.

\bibitem{3DGS}
B.~Kerbl, G.~Kopanas, T.~Leimk{\"u}hler, and G.~Drettakis, ``3d gaussian splatting for real-time radiance field rendering,'' \emph{ACM Transactions on Graphics}, vol.~42, no.~4, pp. 1--14, 2023.

\bibitem{DTAM}
R.~A. Newcombe, S.~J. Lovegrove, and A.~J. Davison, ``Dtam: Dense tracking and mapping in real-time,'' in \emph{2011 international conference on computer vision}.\hskip 1em plus 0.5em minus 0.4em\relax IEEE, 2011, pp. 2320--2327.

\bibitem{LSD-SLAM}
J.~Engel, T.~Sch{\"o}ps, and D.~Cremers, ``Lsd-slam: Large-scale direct monocular slam,'' in \emph{European conference on computer vision}.\hskip 1em plus 0.5em minus 0.4em\relax Springer, 2014, pp. 834--849.

\bibitem{DROID-SLAM}
Z.~Teed and J.~Deng, ``Droid-slam: Deep visual slam for monocular, stereo, and rgb-d cameras,'' \emph{Advances in neural information processing systems}, vol.~34, pp. 16\,558--16\,569, 2021.

\bibitem{TANDEM}
R.~Craig and R.~C. Beavis, ``Tandem: matching proteins with tandem mass spectra,'' \emph{Bioinformatics}, vol.~20, no.~9, pp. 1466--1467, 2004.

\bibitem{codemapping}
H.~Matsuki, R.~Scona, J.~Czarnowski, and A.~J. Davison, ``Codemapping: Real-time dense mapping for sparse slam using compact scene representations,'' \emph{IEEE Robotics and Automation Letters}, vol.~6, no.~4, pp. 7105--7112, 2021.

\bibitem{probabilistic}
A.~Rosinol, J.~J. Leonard, and L.~Carlone, ``Probabilistic volumetric fusion for dense monocular slam,'' in \emph{Proceedings of the IEEE/CVF Winter Conference on Applications of Computer Vision}, 2023, pp. 3097--3105.

\bibitem{imap}
E.~Sucar, S.~Liu, J.~Ortiz, and A.~J. Davison, ``imap: Implicit mapping and positioning in real-time,'' in \emph{Proceedings of the IEEE/CVF International Conference on Computer Vision}, 2021, pp. 6229--6238.

\bibitem{nice-slam}
Z.~Zhu, S.~Peng, V.~Larsson, W.~Xu, H.~Bao, Z.~Cui, M.~R. Oswald, and M.~Pollefeys, ``Nice-slam: Neural implicit scalable encoding for slam,'' in \emph{Proceedings of the IEEE/CVF Conference on Computer Vision and Pattern Recognition}, 2022, pp. 12\,786--12\,796.

\bibitem{go-slam}
Y.~Zhang, F.~Tosi, S.~Mattoccia, and M.~Poggi, ``Go-slam: Global optimization for consistent 3d instant reconstruction,'' in \emph{Proceedings of the IEEE/CVF International Conference on Computer Vision}, 2023, pp. 3727--3737.

\bibitem{loopy-slam}
L.~Liso, E.~Sandstr{\"o}m, V.~Yugay, L.~Van~Gool, and M.~R. Oswald, ``Loopy-slam: Dense neural slam with loop closures,'' \emph{arXiv preprint arXiv:2402.09944}, 2024.

\bibitem{plg-slam}
T.~Deng, G.~Shen, T.~Qin, J.~Wang, W.~Zhao, J.~Wang, D.~Wang, and W.~Chen, ``Plgslam: Progressive neural scene represenation with local to global bundle adjustment,'' in \emph{Proceedings of the IEEE/CVF Conference on Computer Vision and Pattern Recognition}, 2024, pp. 19\,657--19\,666.

\bibitem{splatam}
N.~Keetha, J.~Karhade, K.~M. Jatavallabhula, G.~Yang, S.~Scherer, D.~Ramanan, and J.~Luiten, ``Splatam: Splat, track \& map 3d gaussians for dense rgb-d slam,'' \emph{arXiv preprint arXiv:2312.02126}, 2023.

\bibitem{gs-slam}
C.~Yan, D.~Qu, D.~Xu, B.~Zhao, Z.~Wang, D.~Wang, and X.~Li, ``Gs-slam: Dense visual slam with 3d gaussian splatting,'' in \emph{Proceedings of the IEEE/CVF Conference on Computer Vision and Pattern Recognition}, 2024, pp. 19\,595--19\,604.

\bibitem{compact-slam}
T.~Deng, Y.~Chen, L.~Zhang, J.~Yang, S.~Yuan, D.~Wang, and W.~Chen, ``Compact 3d gaussian splatting for dense visual slam,'' \emph{arXiv preprint arXiv:2403.11247}, 2024.

\bibitem{ngm-slam}
M.~Li, J.~Huang, L.~Sun, A.~X. Tian, T.~Deng, and H.~Wang, ``Ngm-slam: Gaussian splatting slam with radiance field submap,'' \emph{arXiv preprint arXiv:2405.05702}, 2024.

\bibitem{MonoGS}
H.~Matsuki, R.~Murai, P.~H. Kelly, and A.~J. Davison, ``Gaussian splatting slam,'' \emph{arXiv preprint arXiv:2312.06741}, 2023.

\bibitem{photo-slam}
H.~Huang, L.~Li, H.~Cheng, and S.-K. Yeung, ``Photo-slam: Real-time simultaneous localization and photorealistic mapping for monocular stereo and rgb-d cameras,'' in \emph{Proceedings of the IEEE/CVF Conference on Computer Vision and Pattern Recognition}, 2024, pp. 21\,584--21\,593.

\bibitem{DPVO}
Z.~Teed, L.~Lipson, and J.~Deng, ``Deep patch visual odometry,'' \emph{Advances in Neural Information Processing Systems}, vol.~36, 2024.

\bibitem{COLMAP-Free}
Y.~Fu, S.~Liu, A.~Kulkarni, J.~Kautz, A.~A. Efros, and X.~Wang, ``Colmap-free 3d gaussian splatting,'' in \emph{Proceedings of the IEEE/CVF Conference on Computer Vision and Pattern Recognition (CVPR)}, June 2024, pp. 20\,796--20\,805.

\bibitem{scannet}
A.~Dai, A.~X. Chang, M.~Savva, M.~Halber, T.~Funkhouser, and M.~Nie{\ss}ner, ``Scannet: Richly-annotated 3d reconstructions of indoor scenes,'' in \emph{Proceedings of the IEEE conference on computer vision and pattern recognition}, 2017, pp. 5828--5839.

\bibitem{neslam}
T.~Deng, Y.~Wang, H.~Xie, H.~Wang, J.~Wang, D.~Wang, and W.~Chen, ``Neslam: Neural implicit mapping and self-supervised feature tracking with depth completion and denoising,'' \emph{arXiv preprint arXiv:2403.20034}, 2024.

\bibitem{TUM-RGBD}
J.~Sturm, N.~Engelhard, F.~Endres, W.~Burgard, and D.~Cremers, ``A benchmark for the evaluation of rgb-d slam systems,'' in \emph{2012 IEEE/RSJ international conference on intelligent robots and systems}.\hskip 1em plus 0.5em minus 0.4em\relax IEEE, 2012, pp. 573--580.

\bibitem{replica}
J.~Straub, T.~Whelan, L.~Ma, Y.~Chen, E.~Wijmans, S.~Green, J.~J. Engel, R.~Mur-Artal, C.~Ren, S.~Verma, \emph{et~al.}, ``The replica dataset: A digital replica of indoor spaces,'' \emph{arXiv preprint arXiv:1906.05797}, 2019.

\bibitem{icl-nuim}
A.~Handa, T.~Whelan, J.~McDonald, and A.~Davison, ``A benchmark for {RGB-D} visual odometry, {3D} reconstruction and {SLAM},'' in \emph{IEEE Intl. Conf. on Robotics and Automation, ICRA}, Hong Kong, China, May 2014.

\bibitem{DSO}
J.~Engel, V.~Koltun, and D.~Cremers, ``Direct sparse odometry,'' \emph{IEEE transactions on pattern analysis and machine intelligence}, vol.~40, no.~3, pp. 611--625, 2017.

\bibitem{nicer-slam}
Z.~Zhu, S.~Peng, V.~Larsson, Z.~Cui, M.~R. Oswald, A.~Geiger, and M.~Pollefeys, ``Nicer-slam: Neural implicit scene encoding for rgb slam,'' \emph{arXiv preprint arXiv:2302.03594}, 2023.

\bibitem{eslam}
M.~M. Johari, C.~Carta, and F.~Fleuret, ``Eslam: Efficient dense slam system based on hybrid representation of signed distance fields,'' in \emph{Proceedings of the IEEE/CVF Conference on Computer Vision and Pattern Recognition}, 2023, pp. 17\,408--17\,419.

\bibitem{co-slam}
H.~Wang, J.~Wang, and L.~Agapito, ``Co-slam: Joint coordinate and sparse parametric encodings for neural real-time slam,'' in \emph{Proceedings of the IEEE/CVF Conference on Computer Vision and Pattern Recognition}, 2023, pp. 13\,293--13\,302.

\bibitem{vox-fusion}
X.~Yang, H.~Li, H.~Zhai, Y.~Ming, Y.~Liu, and G.~Zhang, ``Vox-fusion: Dense tracking and mapping with voxel-based neural implicit representation,'' in \emph{2022 IEEE International Symposium on Mixed and Augmented Reality (ISMAR)}.\hskip 1em plus 0.5em minus 0.4em\relax IEEE, 2022, pp. 499--507.

\end{thebibliography}

\end{document}